%% file: emnlp2021.tex
\pdfoutput=1

\documentclass[11pt]{article}

\usepackage{emnlp2021}

\usepackage{times}
\usepackage{latexsym}
\usepackage{bbm}
\usepackage{booktabs}
\usepackage{amsmath}
\usepackage{amsfonts}
\usepackage{amssymb}
\usepackage{appendix}
\usepackage{graphicx}
\usepackage{multirow}
\usepackage{bm}
\usepackage{enumitem}
\usepackage{xcolor}
\usepackage{hyperref}
\usepackage{url}
\usepackage{soul}
\usepackage[subtle]{savetrees}
\usepackage[normalem]{ulem}

\usepackage[T1]{fontenc}

\usepackage[utf8]{inputenc}

\usepackage{microtype}

\title{Improving Numerical Reasoning Skills in the Modular Approach for Complex Question Answering on Text}

\author{
  Xiao-Yu Guo \\
  \\
  {\tt } \\
  \And
  Yuan-Fang Li \\
  Faculty of Information Technology, Monash University, Melbourne, Australia \\
  {\tt \{xiaoyu.guo, yuanfang.li, gholamreza.haffari\}@monash.edu} \\
  \And
  Gholamreza Haffari \\
  \\
  {\tt } \\}

\begin{document}
\maketitle
\begin{abstract}
Numerical reasoning skills are essential for complex question answering (CQA) over text.
It requires opertaions including counting, comparison, addition and subtraction. 
A successful approach to CQA on text, Neural Module Networks (NMNs), follows the \emph{programmer-interpreter} paradigm and leverages specialised modules to perform compositional reasoning. 
However, the NMNs framework does not consider the relationship between numbers and entities in both questions and paragraphs.
We propose effective techniques to improve NMNs' numerical reasoning capabilities by making the interpreter question-aware and capturing the relationship between entities and numbers. 
On the same subset of the DROP dataset for CQA on text, experimental results show that our additions outperform the original NMNs by 3.0 points for the overall F1 score.
\end{abstract}

\input{sec1-intro}

\input{sec2-relat}

\input{sec3-model}

\input{sec4-expr}

\input{sec5-conc}

\bibliography{anthology,custom}
\bibliographystyle{acl_natbib}

\input{sec6-app}

\end{document}

%% file: sec1-intro.tex
\section{Introduction}
\label{sec:introduction}
Complex Question Answering (CQA) is a challenging task, requiring a model to perform compositional and numerical reasoning. 
Originally proposed for the visual question answering (VQA) task, Neural Module Networks (NMNs)~\cite{AndreasRDK16-nmns} have recently been adopted to tackle the CQA problem over text~\cite{GuptaLR0020-nmns}. 
The NMNs is an end-to-end differentiable model in the \emph{programmer-interpreter} paradigm.
Briefly, the \emph{programmer} learns to map each question into a program, i.e.\ a sequence of neural modules, and the \emph{interpreter} then ``executes'' the program, operationalized by modules, on the paragraph to yield the answer for different types of complex questions. 
NMNs achieves the best performance on a subset of the challenging DROP dataset~\cite{dua-etal-2019-drop} and is interpertable by nature.

\begin{figure*}[!htb]
    \centering
    \includegraphics[width=1.0\linewidth]{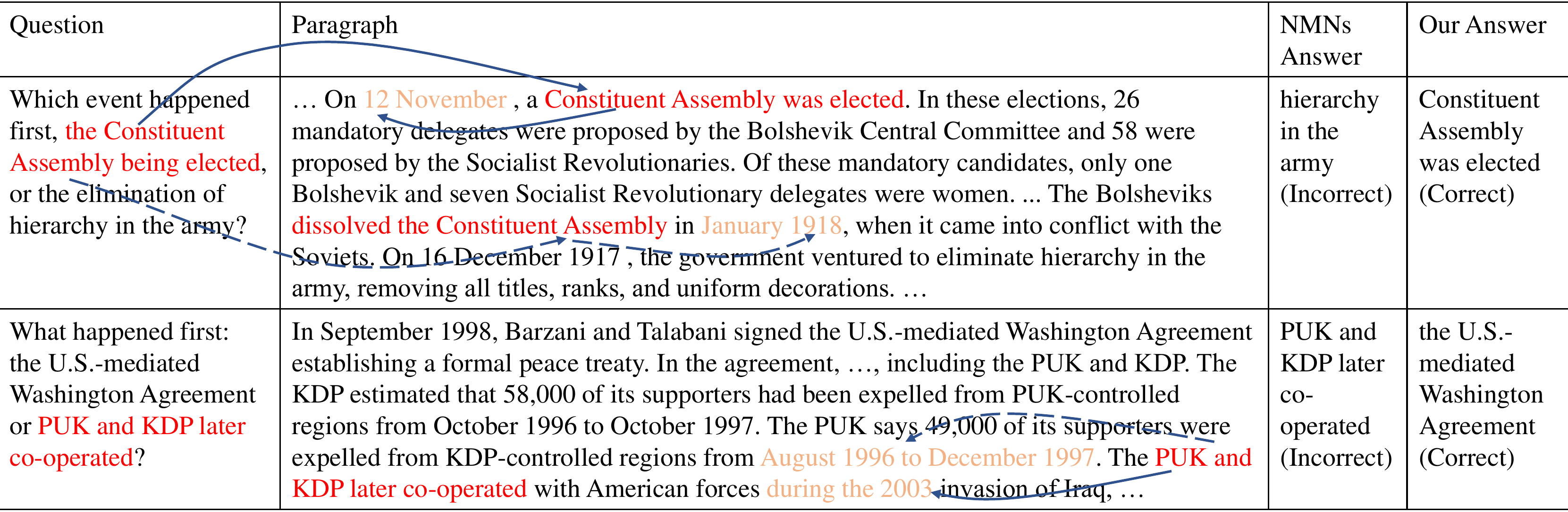}
    \caption{Two examples in the DROP~\cite{dua-etal-2019-drop} dataset that demonstrate the deficienties of NMNs. Tokens pertinent to our discussion are highlighted in red, and their relevant numbers are highlighted in orange. Solid blue lines are predictions of our model, while dotted blue lines show the predictions of NMNs.}
    \label{fig:examples}
\end{figure*}

However, NMNs' performance advantage is not consistent, as it underperforms in some types of questions that require numerical reasoning. 
For instance, for date-compare questions, MTMSN~\cite{HuPHL19-mtmsn} achieves an F1 score of 85.2\footnote{All F1 and EM numbers in this paper are percentages.}, whereas NMNs' performance is 82.6. Similarly, for count questions, the F1 score is 61.6 for MTMSN and 55.7 for NMNs. 
This performance gap stems from two deficiencies of NMNs, which we describe below with the help of two examples in Figure~\ref{fig:examples}. 

Firstly, NMNs' interpreter is \textbf{oblivious to the question when executing number-related modules.}
For executing number-related modules, the interpreter only receives the paragraph as input, but not the question.
Such a lack of direct interactions with the question impairs model performance: the entities in the question, which may also occur in the paragraph, can help locate significant and relevant numbers to produce the final answer.
In the first example in Figure~\ref{fig:examples}, if the interpreter is aware of the correct event mentioned in the question (i.e.\ ``the Constituent Assembly being elected''), it can easily find the same event in the paragraph and further locate its date (``12 November'') precisely. Without this knowledge, the original NMNs found the wrong event (i.e.\ ``dissolved the Constituent Assembly''), thus the wrong date (``January 1918''), leading to an incorrect answer. 


Secondly, NMNs \textbf{disregards the relative positioning of entities and their related numbers} in the paragraph.
Although NMNs can learn separate distributions over numbers extracted from a paragraph, it does not have an effective mechanism to identify the number that \textbf{connects} to a given entity.
Such an ability to recognise \textbf{the association among numbers and entities} is vital for learning numerical reasoning skills: the operation between numbers is meaningful only when they refer to the same entity or the same type of entities.
The second example in Figure~\ref{fig:examples} illustrates the positioning of entities and their related numbers.
With only a constraint on a window around an entity, the NMNs' interpreter tends to identify the nearest number as the related one to a given entity (``August 1996 to December 1997'' for entity ``PUK and KDP later co-operated''), resulting in wrong predictions. 

We propose three simple and effective mechanisms to improve NMNs' numerical reasoning capabilities. 
Firstly, we improve the interpreter to make it question-aware.
By explicitly conditioning the execution on the question, the interpreter can exploit the information contained in the question. 
Secondly, we propose an intuitive constraint to better relate numbers and their corresponding entities in the paragraph. 
Finally, we strengthen the auxiliary loss to increase attention values of entities in closer vicinity within a sentence.
Experimental results show that our modifications significantly improve NMNs' numerical reasoning performance by up to 3.0 absolute F1 points. 
With minor modification, these mechanisms are simple enough to be applied to other modular approaches.

%% file: sec2-relat.tex
\section{Related Work}
\label{sec:realted work}

\noindent\textbf{Complex Question Answering}
focuses on questions that require capabilities beyond multi-hop reasoning. These capabilities include numerical, logical and discrete reasoning.
A number of neural models were recently proposed to address the CQA task, such as BiDAF~\cite{SeoKFH17-bidaf}, QANet~\cite{YuDLZ00L18-qanet}, NMNs~\cite{GuptaLR0020-nmns} and NumNet~\cite{ran-etal-2019-numnet}, which achieved high performance on benchmark datasets such as DROP~\cite{dua-etal-2019-drop}.

\noindent\textbf{Numerical Reasoning}
is an essential capability for the CQA task, which is a challenging problem since the numbers and computation procedures are separately extracted and generated from raw text.
\citet{dua-etal-2019-drop} modified the output layer of QANet~\cite{YuDLZ00L18-qanet} and proposed a number-aware model NAQANet that can deal with numerical questions for which the answer cannot be directly extracted from the paragraph.
In addition to NAQANet, NumNet~\cite{ran-etal-2019-numnet} leveraged Graph Neural Network (GNN) to design a number-aware deep learning model.
Also leveraging GNN, \citet{chen-etal-2020-qdgat} distinguished number types more precisely by adding the connection with entities and obtained better performance.
\citet{ChenLYZSL20-nerd} searched possible programs exhaustively based on answer numbers and employed these programs as weak supervision to train the whole model.
Using dependency parsing of questions, \citet{abs-2101-11802-wsnmns} focused on the numerical part and obtained excellent results on different kinds of numerical reasoning questions.

\noindent\textbf{Neural Module Networks (NMNs)}~\cite{GuptaLR0020-nmns} 
\label{subsec:nmns}
adopts the \emph{programmer-interpreter} paradigm and is a fully end-to-end differentiable model, in which the programmer (responsible for composing programs) and the interpreter (responsible for \emph{soft} execution) are jointly learned. 
Specialised modules, such as \textit{find} and \textit{find-num}, are predefined to perform different types of reasoning over text and numbers.
Compared with those techniques that employ GNNs~\cite{ran-etal-2019-numnet,YuDLZ00L18-qanet}, NMNs is highly interpretable while achieving competitive performance.
More details can be found in Appendix \ref{app:nmns}.

%% file: sec3-model.tex
\section{Proposed Model}
\label{sec:proposed model}
In this section, we will discuss the deficiencies of NMNs described in Section~\ref{sec:introduction} and propose three techniques to overcome these problems. 
Considering the importance of questions while executing programs, we incorporate a question-to-paragraph alignment matrix to form a question-aware interpreter in Section \ref{subsec:alignment}.
In Section \ref{subsec:limitation}, the correspondence between numbers and their related entities is enhanced with a simple and effective constraint on number-related modules. 
In Section \ref{subsec:auxiliary}, we strengthen the auxiliary loss function in NMNs to further concentrate attention in the same sentence.

\subsection{Question-aware Interpreter}
\label{subsec:alignment}
The interpreter in the NMNs framework is responsible for executing specialised modules given the context (i.e.\ paragraph). For number-related modules such as \emph{``find-num''}, the question is not taken into account, which limits NMNs' performance on numerical reasoning, as information in the question is not taken into account. 
As an example, let us take a clear look at the \textit{``find-num''} module in NMNs.

\noindent\textbf{find-num}($\mathcal{P}$) $\rightarrow \mathcal{T}$\footnote{We follow \citet{GuptaLR0020-nmns} and use same variables, annotations in equations for consistency.}. This module takes as input the distribution over paragraph tokens, and produces output an distribution over the numbers:

\vspace{-8pt}
\begin{align}
\mathbf{S^n_{ij}} &= \mathbf{P_i}^T \mathbf{W_n} \mathbf{P_{n_j}}, \label{eq:sim}\\ 
\mathbf{A^n_i} &= softmax(\mathbf{S^n_i}), \label{eq:attention} \\
\mathcal{T} &= \sum_\mathbf{i} \mathcal{P}_\mathbf{i} \cdot \mathbf{A^n_i},  \label{eq:number distribution}
\end{align}
where input $\mathcal{P}$ and output $\mathcal{T}$ are distributions over paragraph tokens and numbers respectively, 
$\mathbf{P}$ is the paragraph token representations, 
$\mathbf{i}$ is the index of the $i^{th}$ paragraph token, 
$\mathbf{n_j}$ is the index of the $j^{th}$ number token, 
and $\mathbf{W_n}$ is a learnable matrix.
Note that when computing the similarity matrix between the paragraph token $\mathbf{P_i}$ and the number token $\mathbf{P_{n_j}}$ in Equation~\ref{eq:sim}, there is no interaction with the question. 

When the correct number types or related entities can be easily found in the question, incorporating the question in \textit{``find-num''} can help narrow down the search of numbers in the paragraph.
The first example in Figure \ref{fig:examples} shows that the NMNs fails to locate the correct number as the wrong event is recognized, without interacting with the question.

Inspired by this idea, we propose the \emph{question-to-paragraph alignment} modification to number-related modules.
Specifically, the definition of \textit{``find-num''} is modified as follows:

\noindent\textbf{find-num}($\mathcal{P}$, $\mathcal{Q}$) $\rightarrow \mathcal{T}^n$, where the additional input $\mathcal{Q}$ obtained from the programmer represents the distribution over question tokens, and the new output is represented by $\mathcal{T}^n$.
Additional computational steps (Equation~\ref{eq:s_prime} to \ref{eq:tn} below) are \textbf{added} after Equation \ref{eq:number distribution}:

\begin{align}
\mathbf{S^{n'}_{kj}} &= \mathbf{Q_k}^T \mathbf{W_n} \mathbf{P_{n_j}},\label{eq:s_prime}\\
\mathbf{A^{n'}_k} &= softmax(\mathbf{S^{n'}_k}),\\
\mathcal{T}' &= \sum_\mathbf{k} \mathcal{Q}_\mathbf{k} \cdot \mathbf{A^{n'}_k},\label{eq:t_prime}\\
\mathcal{T}^n &= \lambda \cdot \mathcal{T} + (1 - \lambda) \cdot \mathcal{T}'\label{eq:tn},
\end{align}
where $\mathbf{Q}$ is the question token representations and $\mathbf{k}$ is the index of the $k^{th}$ question token.

As can be seen from the above equations, the input of the improved \textit{``find-num''} module is extended to include not only paragraph but also question token distributions instead of only the paragraph.
More precisely, $\mathcal{T}'$ is another alignment matrix between all question tokens and number tokens, using the same form of Bi-linear attention computation as $\mathcal{T}$. 

Finally, the new distribution $\mathcal{T}^n$ is produced by the weighted sum of $\mathcal{T}$ and $\mathcal{T}'$ with an additional hyper-parameters $\lambda$.
Here we fix $\lambda = 0.5$ so that NMNs treats the paragraph and the question equally.
Other number-related modules are also revised in a similar way, e.g. \textit{``find-date''}, \textit{``compare-num-lt-than''}, \textit{``find-max-num''}.

\subsection{Number-Entity Positional Constraint}
\label{subsec:limitation}
It is highly likely for a paragraph to contain multiple numbers and entities, as shown in  Figure~\ref{fig:examples}. For such paragraphs, the original NMNs allows all numbers to interact with all entities in the computation of number-related modules such as \textit{``find-num''}. 
This is detrimental to performance as, intuitively, a number far away from an entity is less likely to be related to the entity. 
As the second example in Figure \ref{fig:examples} shows, NMNs connects ``December 1997'' to the entity ``PUK and KDP'' since ``2003'' is far away from it, resulting in wrong predictions eventually.

To tackle this issue, we add another computational component, the relation matrix $\mathbf{U^n}$, into number-related modules.
Taking the \textit{``find-num''} module as an example, the following step is added before Equation \ref{eq:attention} when computing $\mathbf{S^n_{ij}}$:
\begin{align}
\mathbf{S^n_{ij}} &= \mathbf{U^n_{ij}} \circ \mathbf{S^n_{ij}},
\end{align}
where $\circ$ is element-wise multiplication. 
In the above equation, the value of $\mathbf{S^n_{ij}}$ is updated with the relation matrix $\mathbf{U^n}$, which constrains the relationship between the $i^{th}$ paragraph token and $j^{th}$ number token. 
More specifically, let $s_t$ be the token index set for the $t^{th}$ sentence in the paragraph.
Thus, if both the $i^{th}$ paragraph token and the $j^{th}$ number token belong to the same sentence, element $\mathbf{U^n_{ij}}$, in row $i$ and column $j$, is set to 1, otherwise 0:
\begin{align}
\begin{split}
\mathbf{U^n_{ij}} = \left \{
\begin{array}{ll}
    1,              & (\mathbf{i} \in s_t) \wedge (\mathbf{n_j} \in s_t) \\
    0,              & otherwise
\end{array}
\right.
\end{split}
\end{align}

By adding this matrix, the module only keeps the attention values of tokens in close vicinity within a sentence, and learns to find the related numbers that directly interact with entities.
Similarly, this relation matrix $\mathbf{U^n}$ is also applied to other number-related modules to improve performance.

\subsection{Auxiliary Loss Function}
\label{subsec:auxiliary}
\citet{GuptaLR0020-nmns} employed an auxiliary loss to constrain the relative positioning of output tokens with respect to input tokens in the  \textit{``find-num''}, \textit{``find-date''} and \textit{``relocate''} modules.
For instance, the auxiliary loss for the \textit{``find-num''} module is as follows:
\vspace{-4pt}
\begin{equation}
    H^{n}_{loss} = - \sum_{\bf i=1}^{m} \log (\sum_{\bf j=0}^{N_t} \mathbbmss{1}_{\bf n_j \in [i\pm W]} \mathbf{A^n_{ij}}),
\vspace{-4pt}
\end{equation}
where $\mathbf{A^n_{ij}}$ is from Equation \ref{eq:attention}. 
The loss enables the model to concentrate the attention mass of output tokens within a window of size $\mathbf{W}$ (e.g.\ $\mathbf{W} = 10$).

However, these loss functions still allow irrelevant numbers to have spuriously high attention values. 
Taking the second line in Figure \ref{fig:examples} as an example, based on the loss computation procedures, the number ``December 1997'' will be also ``found'' and connected to the entity ``PUK and KDP'' in NMNs.
Obviously, this irrelevant year information should not be taken into consideration.
Therefore, we propose to strengthen the auxiliary loss to further concentrate attention mass to those tokens within the same sentence:
\vspace{-4pt}
\begin{align}
    H^{n}_{loss} = - \sum_{\bf i=1}^{m} \log (\sum_{\bf j=0}^{N_t} \mathbbmss{1}_{(\mathbf{n_j} \in s_t) \wedge (\mathbf{i} \in s_t)} \mathbf{A^n_{ij}}),
\vspace{-4pt}
\end{align}
where the $s_t$ is the token index set for the $t^{th}$ sentence in the paragraph.
In this way, the year ``2003'' is the only consideration for the previous example.

%% file: sec4-expr.tex
\section{Experiments}
\label{sec:experiments}

\subsection{Dataset and Settings}
We evaluate model performance on the same subset of the DROP dataset used by the original NMNs~\cite{GuptaLR0020-nmns}, which contains approx.\ 19,500 QA pairs for training, 440 for validation and 1,700 for testing. 
The training procedures and hyper-parameter settings are the same as the original NMNs~\cite{GuptaLR0020-nmns}.
We report F1 and Exact Match (EM) scores following the literature \cite{dua-etal-2019-drop,GuptaLR0020-nmns}. 

\subsection{Results}

Table \ref{table:numerical reasoning} shows the main results, where  ``original'' represents the performance of the original NMNs~\cite{GuptaLR0020-nmns}. Row 4, ``+qai+nepc+aux'', is our full model, which includes the question-aware interpreter (+qai), the number-entity positional constraint (+nepc), and the improved auxiliary loss (+aux). 
It can be observed that compared to ``original'', our full model achieves significantly higher performance with F1 of 80.4 and EM of 76.6, representing an increase of 3.0 and 2.6 absolute points respectively. Besides, our significant test shows $p \leq 0.01$.

\begin{table}[hbt]
\centering
\begin{tabular}{lcc}
\toprule 
\bf Methods & \bf F1 & \bf EM \\ \midrule
original\cite{GuptaLR0020-nmns} & 77.4 & 74.0 \\ \midrule
ours & & \\
\hspace{3mm}+qai & 79.0 & 74.9 \\
\hspace{3mm}+qai+nepc & 79.9 & 76.0 \\
\hspace{3mm}+qai+nepc+aux & \textbf{80.4} & \textbf{76.6} \\
\bottomrule
\end{tabular}
\caption{\label{table:numerical reasoning} Comparison between different models.}
\vspace{-4pt}
\end{table}

We also conduct an ablation study to discuss the contribution of individual technique. The second line, ``+qai'', is the results with the question-aware interpreter employed only. For this variant, the F1 and EM scores improve on the original baseline by 1.6 and 0.9 points respectively. 
With the addition of the number-entity positional constraint, ``+nepc'', results show an improvement of 2.5 and 2.0 points for F1 and EM when comparing with ``original''.
These results show that all of the three techniques are effective in improving numerical reasoning skills for NMNs.


We also report performance by subsets of different question types in Table \ref{table:question types}.
Except for the number-compare type, our model improves on the original NMNs across all other types of questions significantly, by at least 3.2 absolute points for F1.
In addition, our model outperforms aforementioned MTMSN~\cite{HuPHL19-mtmsn} on all question types as well.

\begin{table}[htb]
\centering
\begin{tabular}{lccc}
\toprule Question Type & MTMSN & original & ours \\ \midrule
date-compare & 85.2 & 82.6 & \bf 86.0 \\ 
date-difference & 72.5 & 75.4 & \bf 78.6 \\
number-compare & 85.1 & \bf 92.7 & 90.1 \\
extract-number & 80.7 & 86.1 & \bf 90.1 \\
count & 61.6 & 55.7 & \bf 61.8 \\
extract-argument & 66.6 & 69.7 & \bf 73.2 \\
\bottomrule
\end{tabular}
\caption{\label{table:question types} Performance (F1) by question types.}
\vspace{-8pt}
\end{table}

%% file: sec5-conc.tex
\section{Conclusion}
\label{sec:conclusion}

Neural Moudule Networks (NMNs) represent an interpretable state-of-the-art approach to complex question answering over text. 
In this paper, we further improve NMNs' numerical reasoning capabilities, by making the interpreter question-aware and placing stronger constraints on the relative positioning of entities and their related numbers. 
Experimental results show that our approach significantly improves NMNs' numerical reasoning ability, with an increase in F1 of 3.0 absolute points.

%% file: sec6-app.tex
\clearpage
\newpage
\appendix
\section{NMNs model overview}
\label{app:nmns}
\begin{figure}[!htb]
    \centering
    \includegraphics[width=1.0\linewidth]{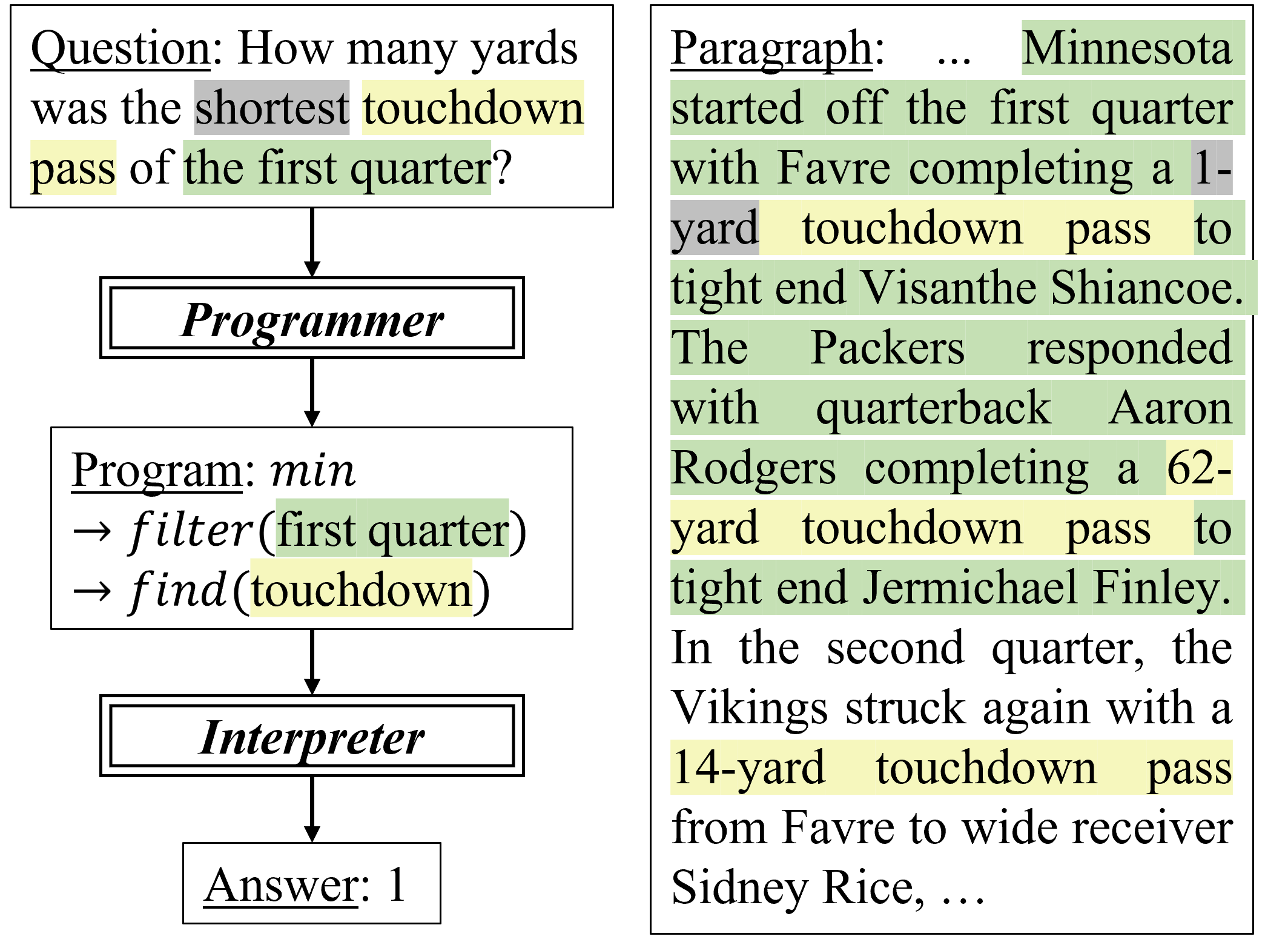}
    \caption{Architecture of the NMNs model.}
    \label{fig:nmns}
\end{figure}

In order to solve the complex question answering problem, \citet{GuptaLR0020-nmns} proposed a Neural Module Networks (NMNs) model. Consisting of a programmer and an interpreter, NMNs can be more interpretable as shown in Figure \ref{fig:nmns}. 

As Figure \ref{fig:nmns} shows, NMNs takes the question and the paragraph as inputs. 
The programmer firstly maps the question into corresponding ``discrete'' modules in order. 
Then, the interpreter executes these generated modules against the corresponding paragraph to produce the final answer. 
Moreover, all modules are differentiable so that the whole NMNs can be trained in an end-to-end way.
\section{Settings for Experiments}
\label{app:settings}
We mainly use PyTorch and AllenNLP deep learning platforms to implement our model.
After 40-epoch training on Ubuntu 16.04 with one V100 GPU Card (16GB memory), it takes around 24 hours to converge. And all reported results are produced based on the saved checkpoint.

\begin{table}[htb]
    \centering
    \begin{tabular}{l|c}
         \hline Name & Value \\ \hline
         batch size & 4 \\
         epochs & 40 \\
         hard em epochs & 5 \\
         learning rate & 1e-5 \\
         drop out rate & 0.2 \\
         max question length & 50 \\
         max paragraph length & 459 \\
         max decode step & 14 \\ \hline
    \end{tabular}
    \caption{Hyper-parameter settings.}
    \label{tab:hyper}
\vspace{-8pt}
\end{table}

For hyper-parameters in our model, we don't conduct experiments on their search trials since we employ the same settings as \citet{GuptaLR0020-nmns} did, which can be found in Table \ref{tab:hyper}. Note that they are also the configuration to obtain the best performance.
For the added parameter $\lambda$ in Equation \ref{eq:tn}, we leverage an empirical value $\lambda = 0.5$ without any fine-tuning. 

Due to the page limitation, we didn’t include more baselines, such as NAQANet~\cite{dua-etal-2019-drop}. 
After running on the same split of DROP dataset, the F1 and EM scores by NAQANet are 62.1\% and 57.9\% respectively, which are substantially lower than our results in Table \ref{table:numerical reasoning}, by over 17\% for both scores.
And we did apply these components in Section \ref{sec:proposed model} to other modules, such as the “extract-argument” module (extracts spans or tokens from paragraphs), and also obtained better results (0.5\% F1 increase). 
Besides, for different question types, their statistics on the test set can be found in Table \ref{tab:statistics}.

\begin{table}[htb]
\centering
\begin{tabular}{lc}
\toprule Question Type & Percentage \\ \midrule
date-compare & 18.6\%  \\ 
date-difference & 17.9\% \\
number-compare & 19.3\% \\
extract-number & 13.5\% \\
count & 17.6\%  \\
extract-argument & 12.8\% \\
\bottomrule
\end{tabular}
\caption{Percentage by question types.}
\label{tab:statistics} 
\vspace{-8pt}
\end{table}

Current NMNs~\cite{GuptaLR0020-nmns} does not support other arithmetic datasets, since some arithmetic operations, including addition, are not supported. 
Extending related arithmetic modules is one of our future work, based on which the NMNs could be trained on other datsets.